# Practical Coreset Constructions for Machine Learning

Olivier Bachem⋆, Mario Lucic⋆, Andreas Krause
Department of Computer Science, ETH Zurich
⋆ These authors contributed equally.

# Contents

# 1
## Introduction

Over the last years, the world has witnessed the emergence of data sets of an unprecedented size across different scientific disciplines. The large volume of such data sets presents new challenges as gathering, storing, and analyzing them becomes expensive. In the context of millions or even billions of data points, existing proven algorithms "suddenly" become computationally infeasible while data sets may not fit on single machines anymore but must be stored on clusters of machines.

As a consequence, new algorithms are required to scale to this *massive data* setting. While one could focus on single machine learning problems and come up with endless new algorithms, we focus on a more general approach: we investigate *coresets* — succinct, small summaries of large data sets — so that solutions found on the summary are *provably competitive* with solution found on the full data set.

Coresets were originally studied in the context of computational geometry and older approaches often relied on computationally expensive methods such as exponential grids. Recently, a new sampling based approach has led to a variety of practical coreset constructions for machine learning problems (see Section 3 for an overview). First, one identifies how important the different data points are with regards to the objec-tive and the optimal solution. This can often be done very efficiently, i.e., in linear time, and in a distributed setting. Then, this importance information is used to construct a coreset by taking a weighted subsample of the data based on importance sampling. Finally, one uses existing algorithms to solve the machine learning problem on the core set while retaining strong theoretical guarantees on the solution quality.

The size of the generated coresets is usually sublinear in if not independent of the original number of data points. Hence, even computationally intensive inference algorithms with superlinear complexity may be used to solve the original machine learning problem on the small coreset.

In this work, we aim to provide an overview over the state-of-the-art in coreset construction for machine learning. In Section 2, we present both the intuition behind and a theoretically sound framework to construct coresets for general problems and explain it at the example of $k$-means clustering. In Section 3, we then summarize existing coreset construction for a variety of machine learning problems such as maximum likelihood estimation of mixture models, Bayesian non-parametric models, principal component analysis, regression and general empirical risk minimization.

# 2

# Practical Coreset Constructions via Importance Sampling

The goal of this chapter is to provide a gentle introduction to coreset construction via importance sampling. We start with the intuition behind the approach and gradually develop a formal coreset construction framework culminating in the main theorem in Section 2.5. While the approach itself is broadly applicable, we will illustrate it through the use of *k*-means clustering as a prototypical machine learning problem.

## 2.1 What are coresets?

We introduced *coresets* as small, weighted summaries of large data sets such that solutions found on the summary itself are *provably competitive* with solutions found on the full data set. While this definition conveys the key idea behind coresets, it is also vague as the notion of *provably competitive* depends on the underlying machine learning problems. First, we hence introduce a formal notion of coresets for a wide range of machine learning problems.

A common — if not the standard — approach in machine learning is to formulate learning problems as optimization problems. Let $\mathcal{X}$ be a weighted data set where $x \in \mathcal{X}$ is a data point and $\mu_{\mathcal{X}}(x)$ denotes its corresponding (non-negative) weight. Given the data set $\mathcal{X}$ and a space of possible solutions $\mathcal{Q}$, one aims to find a solution $Q \in \mathcal{Q}$ that minimizes a *cost function*, *i.e.*, $\text{cost}(\mathcal{X}, Q)$. In this work, we focus on cost functions $\text{cost}(\mathcal{X}, Q)$ that are additively decomposable into non-negative functions $f_Q(x)$, *i.e.*,

$$\text{cost}(\mathcal{X}, Q) := \sum_{x \in \mathcal{X}} \mu_{\mathcal{X}}(x) f_Q(x). \tag{2.1}$$

Instances of such machine learning problems include support vector machines, logistic regression, linear regression and $k$-means clustering. For example, the goal of $k$-means clustering is to find a set of $k$ cluster centers in $\mathbb{R}^d$ minimizing the *quantization error*

$$\text{cost}(\mathcal{X}, Q) = \sum_{x \in \mathcal{X}} \mu_{\mathcal{X}}(x) f_Q(x) = \sum_{x \in \mathcal{X}} \mu_{\mathcal{X}}(x) \min_{q \in Q} ||x - q||_2^2$$

where $f_Q(x) = \min_{q \in Q} ||x - q||_2^2$ is the squared distance from each point $x \in \mathcal{X}$ to the closest cluster center in $Q$.

The key idea behind coresets is to approximate the original data set $\mathcal{X}$ by a (potentially) weighted set $\mathcal{C}$. The most popular notion of coresets requires that the set $\mathcal{C}$ provides a $1 \pm \epsilon$ multiplicative approximation of the cost function as specified in Definition 2.1.

**Definition 2.1** (Coreset)**.** Let $\varepsilon > 0$. The weighted set $\mathcal{C}$ is a $\epsilon$-*coreset* of $\mathcal{X}$ if for all $Q \in \mathcal{Q}$

$$|\text{cost}(\mathcal{X}, Q) - \text{cost}(\mathcal{C}, Q)| \leq \varepsilon \, \text{cost}(\mathcal{X}, Q). \tag{2.2}$$

If the guarantee (2.2) holds uniformly for all possible solutions $Q \subset \mathcal{Q}$, $\mathcal{C}$ is called a *strong coreset*. In contrast, if the guarantee holds only for the optimal solution $Q^\star \in \mathcal{Q}$, $\mathcal{C}$ is called a *weak coreset*. Intuitively, we may *query* $\text{cost}(\mathcal{C}, Q)$ to obtain the cost of any solution $Q$ and are guaranteed an answer within $1 \pm \epsilon$ of $\text{cost}(\mathcal{X}, Q)$. As a result, $Q$ is often called the *query* and $\mathcal{Q}$ the *query space* in the coreset literature.

Coresets are a useful tool as the optimal solution found on the coreset is *provably* competitive with the optimal solution on the full data set when evaluated on the full data set.

**Theorem 2.1.** Let $\epsilon \in (0, \frac{1}{3})$ and $\mathcal{C}$ be a $\epsilon$-coreset of $\mathcal{X}$. Denote by $Q_\mathcal{X}^*$ and $Q_\mathcal{C}^*$ the optimal solution for $\mathcal{X}$ and $\mathcal{C}$ respectively. Then,

$$\mathrm{cost}(\mathcal{X}, Q_\mathcal{C}^*) \leq (1 + 3\epsilon)\,\mathrm{cost}(X, Q_\mathcal{X}^*).$$

*Proof.* By the coreset property and since $Q_\mathcal{C}^*$ is optimal on $\mathcal{C}$, we have

$$\mathrm{cost}(\mathcal{X}, Q_\mathcal{C}^*) \leq \frac{1}{1-\epsilon}\,\mathrm{cost}(\mathcal{C}, Q_\mathcal{C}^*) \leq \frac{1}{1-\epsilon}\,\mathrm{cost}(\mathcal{C}, Q_\mathcal{X}^*)$$
$$\leq \frac{1+\epsilon}{1-\epsilon}\,\mathrm{cost}(\mathcal{X}, Q_\mathcal{X}^*) \leq (1 + 3\epsilon)\,\mathrm{cost}(X, Q_\mathcal{X}^*)$$

where the last step holds as

$$\frac{1+\epsilon}{1-\epsilon} = \frac{1-\epsilon+2\epsilon}{1-\epsilon} = 1 + \frac{2\epsilon}{1-\epsilon} \leq 1 + 3\epsilon.$$

□

## 2.2 Naive approaches to constructing coresets

The existence of *coresets* is trivial — the original data set itself is in fact a coreset. The key question is the existence of *small* coresets where the coreset size is sublinear in — if not independent of — the number of data points $n = |\mathcal{X}|$, while at the same time being polynomial in other parameters, in particular the dimension and the desired error.

A naive approach to constructing coresets is based on uniform subsampling of the data. For the sake of simplicity, consider a data set $\mathcal{X}$ with constant weights $\mu_\mathcal{X}(x) = \frac{1}{|\mathcal{X}|}$. For any query $Q \in \mathcal{Q}$, the cost function in (2.1) may be rewritten as

$$\mathrm{cost}(\mathcal{X}, Q) = \sum_{x \in \mathcal{X}} \frac{1}{|\mathcal{X}|} f_Q(x) = \mathbb{E}_x\left[f_Q(x)\right] \qquad (2.3)$$

where $x \in \mathcal{X}$ is drawn uniformly at random. Let the set $\mathcal{C}$ consist of $m$ points sampled uniformly at random from $\mathcal{X}$ and set $\mu_\mathcal{C}(x) = \frac{1}{m}$. By (2.3), $\mathrm{cost}(\mathcal{C}, \mathcal{Q})$ is an unbiased estimator of $\mathrm{cost}(\mathcal{X}, \mathcal{Q})$ for any query $Q \in \mathcal{Q}$ as

$$\mathbb{E}_\mathcal{C}\left[\mathrm{cost}(\mathcal{C}, \mathcal{Q})\right] = \sum_{x \in \mathcal{C}} \frac{1}{m}\mathbb{E}_x\left[f_Q(x)\right] = \mathrm{cost}(\mathcal{X}, \mathcal{Q}). \qquad (2.4)$$

Furthermore, for any given query $Q \in \mathcal{Q}$, the variance is bounded as

$$\begin{aligned} \mathrm{Var}\left[\mathrm{cost}(\mathcal{C}, \mathcal{Q})\right] &= \frac{1}{m} \mathrm{Var}\left[f_Q(x)\right] \leq \frac{1}{m} \mathbb{E}\left[f_Q(x)^2\right] \\ &= \frac{1}{nm} \sum_{x \in \mathcal{X}} f_Q(x)^2 \leq \frac{1}{nm} \left(\sum_{x \in \mathcal{X}} f_Q(x)\right)^2 \\ &= \frac{n}{m} \mathrm{cost}\left(\mathcal{X}, \mathcal{Q}\right)^2 < \infty. \end{aligned} \qquad (2.5)$$

By the law of large numbers, unbiasedness and finite variance implies that, for any given query $Q \in \mathcal{Q}$, $\mathrm{cost}(\mathcal{C}, \mathcal{Q})$ converges almost surely to $\mathrm{cost}(\mathcal{X}, \mathcal{Q})$ as $m \to \infty$. Unfortunately, this convergence can be slow, requiring sample sizes $m$ that are impractically large. In fact, an application of Chebyshev's inequality to (2.4) and (2.5) yields, for any given query $Q \in \mathcal{Q}$,

$$\mathbb{P}\left[|\mathrm{cost}(\mathcal{X}, Q) - \mathrm{cost}(\mathcal{C}, Q)| > \varepsilon \, \mathrm{cost}(\mathcal{X}, Q)\right] \leq \frac{n}{\varepsilon^2 m}.$$

This result implies that at least $m \geq \frac{n}{\varepsilon^2 \delta}$ samples are required to guarantee that the coreset property in (2.2) holds with probability at least $\delta$ for a single query $Q \in \mathcal{Q}$. Clearly, such a result is not useful as instead of sampling $\Omega(n)$ points we could just return the full data set of size $n$.

This result is tight in $n$ as shown by the following example: Let $\mathcal{X}$ be a data set of $n$ points in $\mathbb{R}$ where $n-1$ points are placed at 0 and one point is placed at 1. We consider $k$-means clustering with $k=1$ and the query $Q = \{0\}$. By definition, we have $\mathrm{cost}(\mathcal{X}, Q) = \frac{1}{n}$. If the set $\mathcal{C}$ does not contain 1, then $\mathrm{cost}(\mathcal{C}, Q) = 0$. As a result, (2.2) can only hold if $\mathcal{C}$ contains the single point at 1. By the union bound, the probability that $\mathcal{C}$ does not contain the single point at 1 is $\left(1 - \frac{1}{n}\right)^m \geq 1 - \frac{m}{n}$. Hence, for (2.2) to hold with constant probability the sample size $m$ needs to be at least $\Omega(n)$.

Even for a single query $Q \in \mathcal{Q}$, this simple example shows that a naive approach based on uniform subsampling fails as a single point may dominate the cost of a query. As a consequence, the subsampling scheme may have a very large variance requiring a large number of samples.

## 2.3 Importance sampling & sensitivity

The example in the previous section suggests that a more advanced sampling scheme is required. *Importance sampling* provides a well-established approach to skew the sampling towards *important* points while retaining an unbiased estimate of the cost function.

**Importance sampling.** For any distribution $q(\cdot)$ on $\mathcal{X}$, importance sampling returns a set $\mathcal{C}$ of $m$ points from $\mathcal{X}$ as follows: First, each point $x \in \mathcal{X}$ is sampled with probability $q(x)$. This is repeated until $\mathcal{C}$ consists a total of $m$ points. Then, each sampled point $x \in \mathcal{C}$ is assigned the weight

$$\mu_\mathcal{C}(x) = \frac{\mu_\mathcal{X}(x)}{mq(x)}.$$

By definition, the resulting sampling scheme is unbiased since for each query $Q \in \mathcal{Q}$ by the linearity of expectation

$$\begin{aligned}
\mathbb{E}\left[\text{cost}(\mathcal{C}, Q)\right] &= \mathbb{E}\left[\sum_{x \in \mathcal{C}} \mu_\mathcal{C}(x) f_Q(x)\right] \\
&= \mathbb{E}\left[\sum_{x \in \mathcal{C}} \frac{\mu_\mathcal{X}(x)}{mq(x)} f_Q(x)\right] \\
&= \mathbb{E}\left[\frac{\mu_\mathcal{X}(x)}{q(x)} f_Q(x)\right] \\
&= \sum_{x \in \mathcal{X}} \mu_\mathcal{X}(x) f_Q(x) \\
&= \text{cost}(\mathcal{X}, Q).
\end{aligned} \qquad (2.6)$$

Given unbiasedness, we would like to find a sampling distribution $q(\cdot)$ that minimizes the variance of the estimator. Intuitively, we would like to sample *influential* points — the ones with a potentially high impact on the cost function — more frequently.

**Single query.** To obtain the intuition how such a sampling distribution $q(\cdot)$ should look like, we first consider a single query $Q \in \mathcal{Q}$. In this case, it is easy to derive the sampling distribution $q(\cdot)$ that minimizes the variance of $\text{cost}(\mathcal{C}, Q)$. Since all $x \in \mathcal{C}$ are independently

sampled, we have

$$\begin{aligned}
\text{Var}\left[\text{cost}(\mathcal{C}, Q)\right] &= \frac{1}{m}\text{Var}\left[\frac{\mu_{\mathcal{X}}(x)}{q(x)}f_Q(x)\right] \\
&= \frac{1}{m}\mathbb{E}\left[\left(\frac{\mu_{\mathcal{X}}(x)}{q(x)}f_Q(x)\right)^2\right] - \frac{1}{m}\mathbb{E}\left[\frac{\mu_{\mathcal{X}}(x)}{q(x)}f_Q(x)\right]^2 \\
&= \frac{1}{m}\sum_{x \in \mathcal{X}}\frac{\mu_{\mathcal{X}}(x)^2}{q(x)}f_Q(x)^2 - \frac{1}{m}\text{cost}(\mathcal{X}, Q)^2
\end{aligned}$$

For any query $Q \in \mathcal{Q}$, the variance is minimized if we choose the sampling distribution

$$q_Q(x) = \frac{\mu_{\mathcal{X}}(x)f_Q(x)}{\sum_{x' \in \mathcal{X}}\mu_{\mathcal{X}}(x')f_Q(x')},$$

as $\text{Var}\left[\text{cost}(\mathcal{C}, Q)\right] = 0$ for $q_Q(\cdot)$. Coupled with with (2.6) this implies that the coreset property in (2.2) holds almost surely for the *single* query $Q \in \mathcal{Q}$. While this may seem promising, we are however interested in a single sampling distribution $q(\cdot)$ which guarantees the coreset property in (2.2) *uniformly* for all queries $Q \in \mathcal{Q}$.

**Sensitivity.** Langberg and Schulman [2010] show that it is sufficient to consider the worst-case query $Q \in \mathcal{Q}$ for each point $x \in \mathcal{X}$. They define the notion of *sensitivity* — the worst-case impact of each data point on the objective function.

**Definition 2.2** (Sensitivity). The *sensitivity* of a point $x \in \mathcal{X}$ w.r.t $f : \mathcal{X} \times \mathcal{Q} \to \mathbb{R}_{\geq 0}$ is defined as

$$\sigma(x) = \sup_{Q \in \mathcal{Q}} \frac{f_Q(x)}{\sum_{x' \in \mathcal{X}}\mu_{\mathcal{X}}(x')f_Q(x')}.$$

The *total sensitivity* is defined as $\mathfrak{S} = \sum_{i=1}^{n}\mu_{\mathcal{X}}(x)\sigma(x)$.

Let $s(x)$ be any upper bound on $\sigma(x)$ and define $S = \sum_{x \in \mathcal{X}}\mu_{\mathcal{X}}(x)s(x)$. In Section 2.5 we will show a general approach for computing $s(\cdot)$ using $k$-means as an example. Consider the sensitivity based importance sampling scheme with

$$q(x) = \frac{\mu_{\mathcal{X}}(x)s(x)}{S}.$$

Given $m$ sufficiently large, we claim that this sampling scheme provides valid $\varepsilon$-coresets. In this section, we provide the intuition behind this sampling scheme while deferring the full technical proof to Section 2.5.

For all $Q \in \mathcal{Q}$ and $x \in \mathcal{X}$, define the function

$$\begin{aligned}g_Q(x) &= \frac{\mu_{\mathcal{X}}(x)f_Q(x)}{\text{cost}(\mathcal{X},Q)}\frac{1}{Sq(x)} \\ &= \frac{f_Q(x)}{\sum_{x'\in\mathcal{X}}\mu_{\mathcal{X}}(x')f_Q(x')}\frac{1}{s(x)}.\end{aligned} \quad (2.7)$$

For any $Q \in \mathcal{Q}$ and $x \in \mathcal{X}$, $g_Q(x)$ is bounded in $[0,1]$ by the definition of sensitivity. By Hoeffding's inequality, we thus have for any $Q \in \mathcal{Q}$ and $\varepsilon' > 0$

$$\mathbb{P}\left[\left|\mathbb{E}\left[g_Q(x)\right] - \frac{1}{m}\sum_{x\in\mathcal{C}}g_Q(x)\right| > \varepsilon'\right] \leq 2\exp\left(-2m\varepsilon'^2\right).$$

By definition, $\mathbb{E}\left[g_Q(x)\right] = \frac{1}{S}$ while $\frac{1}{m}\sum_{x\in\mathcal{C}}g_Q(x) = \frac{\text{cost}(\mathcal{C},Q)}{S\,\text{cost}(\mathcal{X},Q)}$. As such, for any $Q \in \mathcal{Q}$

$$\mathbb{P}\left[|\text{cost}(\mathcal{X},Q) - \text{cost}(\mathcal{C},Q)| > \varepsilon' S\, \text{cost}(\mathcal{X},Q)\right] \leq 2\exp\left(-2m\varepsilon'^2\right). \quad (2.8)$$

Hence, the set $\mathcal{C}$ satisfies the coreset property in (2.2) for any *single* query $Q \in \mathcal{Q}$ and $\varepsilon > 0$ with probability at least $1 - \delta$, if we choose

$$m \geq \frac{S^2}{2\varepsilon^2}\log\frac{2}{\delta}.$$

The required number of samples depends quadratically on the total sensitivity $S$. Hence, the tighter the bound on the sensitivity $s(x) \geq \sigma(x)$ is, the less samples are required. If $\mu_{\mathcal{X}}(x) = \frac{1}{n}$ and we choose the trivial bound $s(x) = n$, then we effectively perform uniform subsampling. In that case, it holds that $S = n$ which implies that we need to sample $\Omega(n^2)$ points which useless in practice. Fortunately, for many machine learning problems, it is possible to compute tighter bounds on the sensitivity as demonstrated in the next section.

## 2.4 Bounding the sensitivity

The standard approach to bound the sensitivity $\sigma(\cdot)$ for a given machine learning problem consists of two steps: In a first step, one finds a

---

**Algorithm 1** $D^2$-sampling
---
**Require:** weighted data set $\mathcal{X}$, number of clusters $k$
 1: Sample $x \in \mathcal{X}$ using $\mu_{\mathcal{X}}(\cdot)$ and set $B = \{x\}$.
 2: **for** $i \leftarrow 2, 3, \ldots, k$ **do**
 3:    Sample $x \in \mathcal{X}$ with probability $\frac{\mu_{\mathcal{X}}(x)\,\mathrm{d}\,(x,B)^2}{\sum_{x' \in \mathcal{X}} \mu_{\mathcal{X}}(x')\,\mathrm{d}\,(x',B)^2}$ and add it to $B$.
 4: **return**   $B$

---

rough approximation of the optimal solution in an efficient manner. In a second step, the said solution is used to bound the worst-case impact of all the points $x \in \mathcal{X}$. We illustrate this approach with the example of $k$-means clustering.

**Rough approximation.** For $k$-means clustering, the notion of a *rough approximation* can be formalized as follows: An $(\alpha, \beta)$-bicriteria approximation is a set of $\beta k$ centers $B$ such that

$$\mathrm{cost}(\mathcal{X}, B) \leq \alpha \,\mathrm{cost}(\mathcal{X}, OPT)$$

where $OPT$ is the optimal solution with $k$ centers.

The state-of-the-art approach is based upon $D^2$-sampling as detailed in Algorithm 1 [Arthur and Vassilvitskii, 2007]: For a data set $\mathcal{X}$ with uniform weights, a first cluster center is sampled uniformly at random from $\mathcal{X}$. Then, in each of $k-1$ subsequent iterations, a center is sampled at random from $\mathcal{X}$ such that each point $x \in \mathcal{X}$ is sampled with probability proportional to the squared distance to the closest of the already sampled cluster centers.

Arthur and Vassilvitskii [2007] show that the resulting set of $k$ centers is in expectation already $\mathcal{O}(\log k)$ competitive, *i.e.*,

$$\mathbb{E}\left[\mathrm{cost}(\mathcal{X}, B)\right] \leq 8\left(\log_2 k + 2\right)\mathrm{cost}(\mathcal{X}, OPT).$$

Let $B^*$ be the set of $k$ centers with the smallest quantization error from $\Theta\left(\log \frac{1}{\delta}\right)$ runs of Algorithm 1. By Markov's inequality, with probability at least $1 - \delta$ it holds that

$$\mathrm{cost}(\mathcal{X}, B^*) \leq 16\left(\log_2 k + 2\right)\mathrm{cost}(\mathcal{X}, OPT).$$

Hence, $B^*$ is a $(\alpha, \beta)$-bicriteria approximation for $\alpha = 16\left(\log_2 k + 2\right)$ and $\beta = 1$ with probability at least $1 - \delta$.

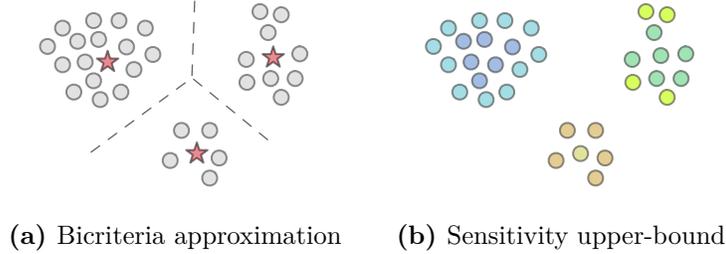

**(a)** Bicriteria approximation  **(b)** Sensitivity upper-bound

**Figure 2.1:** A toy example with three clusters and (a) a bicriteria approximation with 3 centers. (c) Given this approximation we compute the sensitivity of each point using Lemma 2.2 (blue-low probability, red-high). The sampling probabilities are inversely proportional to the cluster size which biases the sampling procedure towards small clusters.

For other bicriteria approximation algorithms for *k*-means clustering offering different tradeoffs in terms of computational complexity and the approximation guarantee we refer the reader to Feldman and Langberg [2011], Makarychev et al. [2016] and Bachem et al. [2016a,b].

**Bounding the sensitivity.** Any such $(\alpha, \beta)$-bicriteria approximation may be used to bound the sensitivity $\sigma(\cdot)$.

**Lemma 2.2.** Let $\mathcal{X} \subset \mathbb{R}^d$ of cardinality $n$ with $\mu_\mathcal{X}(x) = \frac{1}{n}$ and $B \subset \mathbb{R}^d$ be an $(\alpha, \beta)$-bicriteria approximation with respect to the optimal *k*-clustering with squared Euclidean distance. For each point $x \in \mathcal{X}$ denote by $b_x$ the closest cluster center in $B$ and by $\mathcal{X}_x$ the set of all points $x' \in \mathcal{X}$ such that $b_x = b_{x'}$. Then, the sensitivity $\sigma(x)$ of

$$f_Q(x) = \min_{q \in Q} \mathrm{d}\,(x, q)^2$$

is bounded for all $x \in \mathcal{X}$ by the function

$$s(x) = \frac{2\alpha \,\mathrm{d}\,(x, b_x)^2}{\bar{c}_B} + \frac{4\alpha \sum_{x' \in \mathcal{X}_x} \mathrm{d}\,(x', b_x)^2}{|\mathcal{X}_x| \bar{c}_B} + \frac{4n}{|\mathcal{X}_x|}$$

where $\bar{c}_B = \frac{1}{n} \sum_{x' \in \mathcal{X}} \mathrm{d}\,(x', B)^2$. Furthermore,

$$S = \frac{1}{n} \sum_{x \in \mathcal{X}} s(x) = 6\alpha + 4\beta.$$

By Lemma 2.2, selecting the best of $\Theta\left(\log \frac{1}{\delta}\right)$ runs of Algorithm 1 leads to a total sensitivity bound of $\mathcal{O}(k)$. Crucially, $S$ is independent of the number of data points $n$. As shown in Lucic et al. [2016a], this result is tight as there exists a data set $\mathcal{X}$ for which $\mathfrak{S} \in \Omega(k)$.

*Proof of Lemma 2.2.* Consider an arbitrary point $x \in \mathcal{X}$ and an arbitrary query $Q \in \mathcal{Q}$ and define

$$\bar{c}_Q = \frac{1}{n} \sum_{x' \in \mathcal{X}} \mathrm{d}(x', Q) \quad \text{and} \quad \bar{c}_B = \frac{1}{n} \sum_{x' \in \mathcal{X}} \mathrm{d}(x', B).$$

By the double triangle inequality, it holds that

$$\mathrm{d}(x, Q)^2 \leq 2\,\mathrm{d}(x, b_x)^2 + 2\,\mathrm{d}(b_x, Q)^2.$$

Similarly, for all $x' \in \mathcal{X}_x$, it holds that

$$\mathrm{d}(b_x, Q)^2 \leq 2\,\mathrm{d}(x', b_x)^2 + 2\,\mathrm{d}(x', Q)^2.$$

Summing over all points $x \in \mathcal{X}_x$ we obtain

$$\mathrm{d}(b_x, Q)^2 \leq \frac{2}{|\mathcal{X}_x|} \sum_{x' \in \mathcal{X}_x} \left[\mathrm{d}(x', b_x)^2 + \mathrm{d}(x', Q)^2\right].$$

It follows that

$$\frac{f_x(Q)}{\bar{f}_Q} \leq \frac{2}{\bar{c}_Q} \left[\mathrm{d}(x, b_x)^2 + \mathrm{d}(b_x, Q)^2\right]$$

$$\leq \frac{2\,\mathrm{d}(x, b_x)^2}{\bar{c}_Q} + \frac{4 \sum_{x' \in \mathcal{X}_x} \left[\mathrm{d}(x', b_x)^2 + \mathrm{d}(x', Q)^2\right]}{|\mathcal{X}_x| \bar{c}_Q}.$$

By definition of the bicriteria solution, we have both $\bar{c}_Q \geq \frac{1}{\alpha} \bar{c}_B$ and $\bar{c}_Q \geq \frac{1}{|\mathcal{X}_x|} \sum_{x' \in \mathcal{X}_x} \mathrm{d}(x', Q)^2$. Hence

$$\frac{f_x(Q)}{\bar{f}_Q} \leq \frac{2\alpha\,\mathrm{d}(x, b_x)^2}{\bar{c}_B} + \frac{4\alpha \sum_{x' \in \mathcal{X}_x} \mathrm{d}(x', b_x)^2}{|\mathcal{X}_x| \bar{c}_B} + \frac{4n}{|\mathcal{X}_x|} = s(x)$$

is a bound for the sensitivity $\sigma_\mathcal{Q}(x)$ since the choice of both $x \in \mathcal{X}$ and $Q \in \mathcal{Q}$ was arbitrary. We further have

$$S = \frac{1}{n} \sum_{x \in \mathcal{X}} s(x) = 2\alpha + \frac{4}{n} \sum_{x \in \mathcal{X}} \left[\frac{\alpha \sum_{x' \in \mathcal{X}_x} \mathrm{d}(x', b_x)^2}{|\mathcal{X}_x| \bar{c}_B} + \frac{n}{|\mathcal{X}_x|}\right]$$

$$= 6\alpha + 4\beta.$$

which concludes the proof. □

## 2.5 Uniform guarantee for all queries

In Section 2.4, we have shown how to use sensitivity to obtain a set $\mathcal{C}$ that satisfies the coreset property for a *single* query $Q \in \mathcal{Q}$. In this section, we show how to construct an $\varepsilon$-coreset, *i.e.*, a set $\mathcal{C}$ that satisfies the coreset property for *every* query $Q \in \mathcal{Q}$.

Intuitively, we would like to invoke a union bound argument on (2.8) and $\mathcal{Q}$. Since $\mathcal{Q}$ is infinite, directly applying the union bound is rather useless. The key insight is that there exists a (small enough) set of queries $Q^* \subset \mathcal{Q}$ with the following property: if a set $\mathcal{C}$ satisfies the coreset property in 2.2 for each $Q \in Q^*$ with $\varepsilon/2$, then it also satisfies it for each $Q \in \mathcal{Q}$ with $\varepsilon$.

This section requires more advanced tools from computational geometry such as $\varepsilon$-approximations as defined in Li et al. [2001]. As a consequence, the purpose of this section is two-fold: we provide both a formal proof based on (tight) state of the art results and an intuition based on a more accessible example with a looser bound.

**Pseudo-dimension.** For binary functions the classic complexity measure is the *Vapnik-Chervonenkis* (VC) dimension [Vapnik and Chervonenkis, 1971] which is based on the concept of *shattering*: Consider a set of labeled points $\mathcal{X} = \{(x_1, y_1), \ldots, (x_m, y_m)\}$ where $y_i \in \{0, 1\}$. We say that $\mathcal{F} = \{f \mid f : \mathcal{X} \to \{0, 1\}\}$ shatters $\mathcal{X}$ if for all assignments of values $\{0, 1\}$ to labels $y_1, \ldots, y_m$ there exists a function $f \in \mathcal{F}$ which obtains such a labeling. The VC dimension of $\mathcal{F}$ is the largest number of points (arbitrarily arranged) that can be shattered by $\mathcal{F}$. If $\mathcal{F}$ can shatter arbitrary large sets we define the VC dimension to be $\infty$.

In this work we consider a more general setting where the cost functions are not binary and instead map to non-negative reals. We require a generalization of the VC dimension called the *pseudo-dimension*.

**Definition 2.3.** [Pseudo-dimension] Fix a countably infinite domain $\mathcal{X}$. The pseudo-dimension of a set $\mathcal{F}$ of functions from $\mathcal{X}$ to $[0, 1]$, Pdim $(\mathcal{F})$, is the largest $d'$ such there is a sequence $x_1, \ldots, x_{d'}$ of domain elements from $\mathcal{X}$ and a sequence of reals $r_1, \ldots, r_{d'}$ of real thresholds such that for each $b_1, \ldots, b_{d'} \in \{\text{above}, \text{below}\}$, there is an $f \in \mathcal{F}$ such that for all $i = 1, \ldots, d'$, we have $f(x_i) \geq r_i \iff b_i = \text{above}$.

We refer the reader to Anthony and Bartlett [2009] for an extensive discussion and results on the topic.

**Main coreset result.** To obtain coresets, we need to consider the pseudo-dimension of the function family $\mathcal{F} = \{g_Q(\cdot) \mid Q \in \mathcal{Q}\}$ where $g_Q(\cdot)$ are the $[0, 1]$-bounded functions introduced in (2.7). This allows us to show that sensitivity based importance sampling leads to $\varepsilon$-coresets.

**Theorem 2.3** (Coreset size). Let $\varepsilon > 0$ and $\delta \in (0, 1)$. Let $\mathcal{X}$ be a weighted data set, $\mathcal{Q}$ the set of all possible queries and $f_Q(x) : \mathcal{X} \times \mathcal{Q} \to \mathbb{R}_{\geq 0}$ a cost function. Let $s(x) : \mathcal{X} \to \mathbb{R}_{\geq 0}$ denote any upper bound on the sensitivity $\sigma(x)$ and define $S = \sum_{i=1}^{n} \mu_{\mathcal{X}}(x)s(x)$. Let $\mathcal{C}$ be a sample of $m$ points from $\mathcal{X}$ with replacement where each point $x \in \mathcal{X}$ is sampled with probability $q(x) = \frac{\mu_{\mathcal{X}}(x)s(x)}{S}$ and each point $x \in \mathcal{C}$ is assigned the weight $\mu_{\mathcal{C}}(x) = \frac{\mu_{\mathcal{X}}(x)}{mq(x)}$. Let $\text{cost}(\mathcal{X}, Q) = \sum_{x \in \mathcal{X}} \mu_{\mathcal{X}}(x)f_Q(x)$, $\mathcal{F} = \left\{ \frac{\mu_{\mathcal{X}}(\cdot)f_Q(\cdot)}{\text{cost}(\mathcal{X}, Q)Sq(\cdot)} \mid Q \in \mathcal{Q} \right\}$ and $d' = \text{Pdim}(\mathcal{F})$. Then, the set $\mathcal{C}$ is an $\varepsilon$-coreset of $\mathcal{X}$ with probability at least $1 - \delta$ for

$$m \in \Omega\left(\frac{S^2}{\varepsilon^2}\left(d' + \log\frac{1}{\delta}\right)\right).$$

We first provide an informal intuition that is tight up to a factor of $\log\frac{1}{\varepsilon}$ and then provide the full proof based on Li et al. [2001].

**Intuition behind Theorem 2.3.** In principle, for each query $Q \in \mathcal{Q}$, the $m$ sampled points in $\mathcal{C}$ and the corresponding function values $g_Q(x)$ lie in the unit cube $[0, 1]^m$. Intuitively, however, due to the bounded pseudo dimension, the queries $Q \in \mathcal{Q}$ only span a subspace $[0, 1]^{\text{Pdim}(f)}$. Critically, the elements of $\mathcal{Q}$ may be covered up to a $L_1$-distance of $\varepsilon/2$ by a set $\mathcal{Q}^* \subset \mathcal{Q}$ of $\mathcal{O}\left(\varepsilon^{-\text{Pdim}(f)}\right)$ points [Haussler, 1995].

This small set $\mathcal{Q}^*$ possesses the following property: if a set $\mathcal{C}$ satisfies the coreset property in (2.2) for each $Q \in \mathcal{Q}^*$ with $\varepsilon/2$, then it also satisfies it for each $Q \in \mathcal{Q}$ with $\varepsilon$. As a result, we do not need to bound the probability of any of an infinite number of events to occur but we may apply the union bound to $\mathcal{Q}^*$. Let $\mathcal{E}$ denote the bad event i.e.

(2.2) doesn't hold for $\mathcal{C}$. From (2.8), it follows that

$$\mathbb{P}\left[\mathcal{E}\right] = \mathbb{P}\left[\exists Q \in \mathcal{Q} : |\text{cost}(\mathcal{X}, Q) - \text{cost}(\mathcal{C}, Q)| > \varepsilon \, \text{cost}(\mathcal{X}, Q)\right]$$
$$\leq \mathbb{P}\left[\exists Q \in Q^* : |\text{cost}(\mathcal{X}, Q) - \text{cost}(\mathcal{C}, Q)| > \frac{\varepsilon}{2} \text{cost}(\mathcal{X}, Q)\right]$$
$$\leq 2|Q^*|\exp\left(-\frac{m\varepsilon^2}{2S^2}\right).$$

Thus, the set $\mathcal{C}$ is a $\varepsilon$-coreset with probability at least $1 - \delta$ if

$$m \geq \frac{2S^2}{\varepsilon^2}\left(\log|Q^*| + \log\frac{2}{\delta}\right).$$

Given that $|Q^*| \in \mathcal{O}\left(\varepsilon^{-\text{Pdim}(f)}\right)$, we require

$$m \in \Omega\left(\frac{S^2}{\varepsilon^2}\left(\text{Pdim}\,(f)\log\frac{1}{\varepsilon} + \log\frac{1}{\delta}\right)\right)$$

samples to obtain a coreset. This matches the bound in Theorem 2.3 up to the $\log\frac{1}{\varepsilon}$ term.

**Formal proof of Theorem 2.3** The proof is based on the seminal result of Li et al. [2001] which provides a tighter bound based on the *chaining* technique.

**Theorem 2.4** (Li et al. [2001])**.** *Let $\alpha > 0$, $\nu > 0$ and $\delta > 0$. Fix a countably infinite domain $\mathcal{X}$ and let $q(\cdot)$ be any probability distribution over $\mathcal{X}$. Let $\mathcal{F}$ be a set of functions from $\mathcal{X}$ to $[0, 1]$ with $\text{Pdim}\,(\mathcal{F}) = d'$. Denote by $\mathcal{C}$ a sample of $m$ points from $\mathcal{X}$ sampled independently according to $q(\cdot)$. Then, for $m \in \Omega\left(\frac{1}{\alpha^2\nu}(d'\log\frac{1}{\nu} + \log\frac{1}{\delta})\right)$, with probability at least $1 - \delta$ it holds that*

$$\forall f \in \mathcal{F}: \quad \mathrm{d}_\nu\left(\sum_{x \in \mathcal{X}} q(x)f(x), \frac{1}{|\mathcal{C}|}\sum_{x \in \mathcal{C}} f(x)\right) \leq \alpha$$

*where $\mathrm{d}_\nu(a, b) = \frac{|a-b|}{a+b+\nu}$.*

*Proof of Theorem 2.3.* Choose $\nu = \frac{1}{2}$ and $\alpha = \frac{\varepsilon}{3S}$. We apply Theorem 2.4 to the function family $\mathcal{F} = \{g_Q(\cdot) \mid Q \in \mathcal{Q}\}$. This implies that for $m \in \Omega\left(\frac{S^2}{\varepsilon^2}\left(d' + \log\frac{1}{\delta}\right)\right)$, we have with probability at least $1 - \delta$ for all $f \in \mathcal{F}$

$$\frac{\left|\sum_{x \in \mathcal{X}} q(x)f(x) - \frac{1}{|\mathcal{C}|}\sum_{x \in \mathcal{C}} f(x)\right|}{\sum_{x \in \mathcal{X}} q(x)f(x) + \frac{1}{|\mathcal{C}|}\sum_{x \in \mathcal{C}} f(x) + \frac{1}{2}} \leq \alpha.$$

Since both $\sum_{x \in \mathcal{X}} q(x)f(x)$ and $\frac{1}{|\mathcal{C}|}\sum_{x \in \mathcal{C}} f(x)$ are at most one, this implies for all $f \in \mathcal{F}$

$$\left|\sum_{x \in \mathcal{X}} q(x)f(x) - \frac{1}{|\mathcal{C}|}\sum_{x \in \mathcal{C}} f(x)\right| \leq 3\alpha = \frac{\varepsilon}{S}$$

Since $\mathcal{F} = \{g_Q(\cdot) \mid Q \in \mathcal{Q}\}$, we thus have for all $Q \in \mathcal{Q}$

$$\left|\sum_{x \in \mathcal{X}} q(x)g_Q(x) - \frac{1}{|\mathcal{C}|}\sum_{x \in \mathcal{C}} g_Q(x)\right| \leq 3\alpha = \frac{\varepsilon}{S}.$$

By definition of $g_Q$, we have $\sum_{x \in \mathcal{X}} q(x)g_Q(x) = \frac{1}{S}$ while

$$\frac{1}{m}\sum_{x \in \mathcal{C}} g_Q(x) = \frac{\text{cost}(\mathcal{C}, Q)}{S\,\text{cost}(\mathcal{X}, Q)}.$$

Multiplying by $S\,\text{cost}(\mathcal{X}, Q)$ thus implies for all $Q \in \mathcal{Q}$

$$|\text{cost}(\mathcal{X}, Q) - \text{cost}(\mathcal{C}, Q)| \leq \varepsilon\,\text{cost}(\mathcal{X}, Q).$$

which concludes the proof. □

## 2.6 Exemplary coreset construction for $k$-means

In this section, we present a coreset construction for $k$-means based on the presented framework. In Section 2.4, we have derived a sensitivity for $k$-means. To apply Theorem 2.4, we require a bound on the pseudo-dimension of the function family $\mathcal{F} = \{g_Q(\cdot) \mid Q \in \mathcal{Q}\}$ for $k$-means.

**Pseudo-dimension for $k$-means.** By Lemma 1 of Bachem et al. [2017], the pseudo-dimension for the function family $\mathcal{F} = \{g_Q(\cdot) \mid Q \in \mathcal{Q}\}$ is bounded by $\mathcal{O}(dk \log k)$ for $k$-means. The key insight is that the pseudo-dimension of $\mathcal{F}$ may bounded by the VC dimension of a $k$-fold intersections of halfspaces in $\mathcal{O}(d)$-dimensional Euclidean space. While the VC dimension of a halfspace in $R^{\mathcal{O}(d)}$ is $\mathcal{O}(d)$, it not known whether for $d > 3$ $k$-fold intersections of halfspaces are *VC-linear*, i.e., whether the VC dimension is bounded by $\mathcal{O}(dk)$ [Johnson, 2008]. Bachem et al. [2017] use an argument based on Sauer's Lemma to show that an additional $\log k$ factor is sufficient and that the pseudo-dimension for $k$-means may be bounded by $\mathcal{O}(dk \log k)$.

This stands in contrast with previous work in which the bound of $\mathcal{O}(dk)$ is used [Feldman and Langberg, 2011, Balcan et al., 2013, Lucic et al., 2016b]. While all these results rely on Theorem 2.4 by Li et al. [2001], a different definition of the pseudo-dimension is used. In particular, the aforementioned results define the pseudo-dimension to be the the smallest integer $d$ that bounds the number of dichotomies induced by $\mathcal{F}$ on a sample of $m$ points, also called the *growth function*, by $m^d$. However, this bound corresponds to a generalization of the *primal shattering dimension* in classical VC theory to continuous loss functions [Har-Peled, 2011, Matousek, 2009], and is hence *not* the one required by Li et al. [2001].

This discrepancy has the following consequence: A bound on the primal shattering dimension of $d'$ only implies a VC dimension of at most $\mathcal{O}(d' \log d')$ [Har-Peled, 2011]. In the case of $k$-means, the primal shattering dimension bound of $\mathcal{O}(dk)$ [Feldman and Langberg, 2011, Balcan et al., 2013, Lucic et al., 2016b] thus only implies a pseudo-dimension bound of $\mathcal{O}(dk \log dk)$ and leads to correspondingly larger coreset sizes when Theorem 2.4 is applied. In the following, we will thus use the stronger bound of $\mathcal{O}(dk \log k)$ presented above.

**Coreset construction for $k$-means.** The framework and the results for $k$-means presented in this section allow us to derive a coreset construction for $k$-means clustering. First, a rough approximation to the optimal solution is computed using `k-means++` seeding. Then, this solution is used to subsample the data points using importance sampling based on the sensitivity bound derived in Lemma 2.2. The full procedure is detailed in Algorithm 2 and the theoretical guarantee is provided in Theorem 2.5.

**Theorem 2.5.** *Let $\varepsilon \in (0, 1/4), \delta > 0$ and $k \in \mathbb{N}$. Let $\mathcal{X} \subset X$ be a set of points and let $B \subseteq \mathcal{X}$ be the set with the smallest quantization error in terms of d among $\Theta\left(\log \frac{1}{\delta}\right)$ runs of Algorithm 1. Let $C$ be the*

---

**Algorithm 2** Coreset construction

---
**Require:** $\mathcal{X}$, $k$, $B$, $m$
 1: $\alpha \leftarrow 16(\log k + 2)$
 2: **for each** $b_i$ in $B$ **do**
 3: $\quad B_i \leftarrow$ Set of points from $\mathcal{X}$ closest to $b_i$ in terms of d. Ties broken arbitrarily.
 4: $c_\phi \leftarrow \frac{1}{|\mathcal{X}|} \sum_{x' \in \mathcal{X}} \mathrm{d}(x', B)$
 5: **for each** $b_i \in B$ and $x \in B_i$ **do**
 6: $\quad s(x) \leftarrow \frac{\alpha \, \mathrm{d}(x,B)}{c_\phi} + \frac{2\alpha \sum_{x' \in B_i} \mathrm{d}(x',B)}{|B_i| c_\phi} + \frac{4|\mathcal{X}|}{|B_i|}$
 7: **for each** $x \in \mathcal{X}$ **do**
 8: $\quad p(x) \leftarrow s(x) / \sum_{x' \in \mathcal{X}} s(x')$
 9: $\mathcal{C} \leftarrow$ Sample $m$ weighted points from $\mathcal{X}$ where each point $x$ has weight $\frac{1}{m \cdot p(x)}$ and is sampled with probability $p(x)$.
10: **return** $\mathcal{C}$

---

output of Algorithm 2 with

$$m \in \Omega\left(\frac{dk^3 \log k + k^2 \log \frac{1}{\delta}}{\varepsilon^2}\right)$$

Then, with probability at least $1 - \delta$, the set $\mathcal{C}$ is a $(k, \varepsilon)$-coreset of $\mathcal{X}$ for $k$-means.

*Proof.* The proof follows from Theorem 2.3, Lemma 2.2 and the fact that $\mathrm{Pdim}(\mathcal{F}) \in \mathcal{O}(dk \log k)$. □

# 3
# Coresets for Machine Learning

While coresets were developed in the context of computational geometry, we will focus on existing coreset constructions more relevant for the machine learning audience. For an excellent survey on the results from computational geometry consult Agarwal et al. [2005] and Har-Peled [2011]. Furthermore, results stemming from matrix sketching literature are out of the scope of this work and we refer the interested reader to Mahoney [2011], Woodruff et al. [2014] and Phillips [2016].

We review the results on maximum likelihood estimation of mixture models, Bayesian non-parametric models, principal component analysis (PCA), non-negative matrix factorization (NNMF), regression and general empirical risk minimization.

### 3.1  *k*-means, *k*-median, *k*-center and Bregman clustering

Let $\mathcal{X} \subset \mathbb{R}^d$, $Q \subset \mathbb{R}^d$ with $|Q| = k$. The $k$-median and $k$-means cost functions are defined as

$$\text{cost}(\mathcal{X}, Q)_1 = \sum_{x \in \mathcal{X}} \min_{q \in Q} ||x - q||_1,$$

$$\text{cost}(\mathcal{X}, Q)_2 = \sum_{x \in \mathcal{X}} \min_{q \in Q} ||x - q||_2^2.$$

Here we will provide intuition behind the original coreset constructions and introduce the sensitivity based approach discussed in Chapter 2.

**Exponential grids.** The initial coreset constructions relied on first computing a set of $k$ centers which were a constant factor approximation to the optimal clustering. Then, one constructs an exponential grid of $\mathcal{O}(\log n)$ levels around each center and snaps the points to the grid. Har-Peled and Mazumdar [2004] show how this approach leads to coresets of size $\mathcal{O}\left(k\varepsilon^{-d} \log n\right)$ for both $k$-means and $k$-Median. It can be shown that exponential grid approach will always lead to coresets with $\Omega(\log n)$ points. To overcome this issue, Har-Peled and Kushal [2005] instead use a set of $\mathcal{O}\left(1/\varepsilon^{d-1}\right)$ lines and snap the points to the lines instead. Since one can construct coresets for points on a line of size $\mathcal{O}(k/\varepsilon)$, this leads to coresets of size $\mathcal{O}\left(k^2/\varepsilon^d\right)$ for $k$-median clustering, and of size $\mathcal{O}\left(k^3/\varepsilon^{d+1}\right)$ for $k$-means clustering.

**Random sampling.** Chen [2009] proposed coresets with only polynomial dependence on the dimension, for the price of $\log n$. The result is based on a carefully constructed sampling strategy and the seminal result of Haussler [1992]. The data points are assigned to the closest point from the bicriteria approximation. Then, the points are further partitioned according to the distance to the closest center. Then, points are sampled from each partition uniformly at random and their weight is set to be proportional to the size of the partition. This leads to coresets of size $\mathcal{O}(\varepsilon^{-2} k^2 d \log n)$ which can be constructed in $\mathcal{O}(nkd \log(1/\delta))$ time. Based on a similar sampling argument, Feldman et al. [2007] construct a polynomial time approximation scheme based on a coreset of size $\Omega(\varepsilon^{-5} k \log k/\delta)$.

**Importance sampling and sensitivity.** In a breakthrough re-sult Langberg and Schulman [2010] propose a framework approximating non-negative cost functions that can be viewed as expectations by means of random sampling. They show that the critical notion to consider is the *sensitivity*. Based on this novel argument, they construct coresets of size $\mathcal{O}(d^2k^3\varepsilon^{-2})$. Feldman and Langberg [2011] unify previous results and present a general framework for coreset construction: (1) construct a bicriteria approximation for the problem at hand, and (2) bound the sensitivity using the bicriteria approximation. Based on this reasoning they present a coreset for *k*-median of size $\mathcal{O}(kd\varepsilon^{-2})$.

Lucic et al. [2016b] generalize the results to $\mu$-similar Bregman divergences. Here the cost function is given by

$$\text{cost}_\text{H}(\mathcal{X}, Q) = \sum_{x \in \mathcal{X}} \min_{q \in Q} \text{d}_\phi(x, q)$$

where $\text{d}_\phi$ is a Bregman divergence. The family of $\mu$-similar Bregman divergences includes Itakura-Saito distance, KL-divergence, Mahalanobis distance, etc. The authors prove that coresets of size $\mathcal{O}(dk^3\lambda^{-2}\varepsilon^{-2})$ can be constructed in $\mathcal{O}(nkd)$ time. In particular, setting $\text{d}_\phi(x, q) = ||x - q||_2^2$ implies strong coresets of size $\mathcal{O}(dk^3\varepsilon^{-2})$ for *k*-means which is the state-of-the-art.

It is still an open problem whether small coresets exist for the *k*-center problem, i.e.

$$\text{cost}_\infty(\mathcal{X}, Q) = \min_{Q \in \mathcal{Q}} \max_{x \in \mathcal{X}} \text{d}(x, Q),$$

While the problem is NP-Complete, a 2-approximation can be computed in linear time by a greedy strategy Gonzalez [1985]. Har-Peled [2004] show that there exists a multiplicative coreset of size $\mathcal{O}\left(k! \varepsilon^{-dk}\right)$ based on an argument introduced in Agarwal et al. [2002].

**Randomized polynomial-time approximation scheme**. Interestingly, if the size of the strong coreset is independent of the number of data points $n$, one can construct a randomized polynomial-time approximation scheme (PTAS). Consider hard clustering with a $\mu$-similar Bregman divergence. First generate a strong coreset and then consider all possible $k$ partitionings of the coreset points. By the coreset property, it is guaranteed that the centers of the best partitioning are $1 + \varepsilon$ competitive with the optimal solution. Algorithm 3 will output such a solution in $\mathcal{O}\left((nkd + 2^{\text{poly}(\text{kd}/\mu\varepsilon)})\log\frac{1}{\delta}\right)$ time.

---
**Algorithm 3** Randomized PTAS
---
**Require:** $\mathcal{X}$, $k$, $\varepsilon$, d
1: $\mathcal{C} \leftarrow (k, \varepsilon/3)$-coreset for $\mathcal{X}$ with respect to d.
2: $\mathcal{P} \leftarrow$ Centers of all $k$-partitionings of $\mathcal{C}$.
3: $Q^\star \leftarrow \arg\min_{P \in \mathcal{P}} \frac{1}{|\mathcal{C}|} \sum_{(w,c) \in \mathcal{C}} w\, \text{d}(c, P)$
4: **return** $Q^\star$
---

**Using excess data as a computational resource.** Classical learning theory centers around the question of how risk scales with the data set size. From a practical point of view, increasing the data size is a source of computational complexity which typically translates into higher running time. From this perspective, large data is considered a nuisance rather than a resource for achieving lower risk. Using $k$-means as an example, Lucic et al. [2015] show that coresets can be used to strategically summarize the data such that, for a fixed risk level, the running time decreases as the data set size grows. In other words, for a predefined clustering quality, the algorithm outputs a solution faster if more data is available.

## 3.2 Projective clustering, PCA and NNMF

Given a matrix $A \in \mathbb{R}^{n \times d}$ the goal of $(k, j)$-*linear (affine) projective clustering* is to find the minimizer of

$$\text{cost}(A, Q) = \sum_{i=1}^{n} \min_{q \in Q} \text{d}(x_i, q),$$

where $Q \in \mathcal{Q}$ is a set of all $k$ linear (affine) subspaces each of dimension $j$, and $\text{d}(x, q)$ is the squared Euclidean distance from $x$ to its projection onto the subspace $q$. Here the dimension of a subspace $q$ is the number of vectors in any basis of $q$. As such, $(k, 0)$-projective clustering corresponds to $k$-clustering problems including $k$-means and $k$-median, $(k, 1)$-projective clustering corresponds to the $k$-line problem. For general $(k, j)$-projective clustering the linear dependence on $n$ is necessary even for the case $(k = 2, j = 2, d = 3)$ [Har-Peled, 2004]. A particularly interesting case is $(1, j)$-subspace clustering.

**PCA and NNMF as $j$-subspace clustering**. The special case $(1, j)$ of projective clustering (also known as $j$-subspace clustering) has many applications such as principal component analysis and latent semantic analysis. It is well known that the optimal $j$-dimensional subspace is spanned by the top $j$ singular vectors of $A$ and can be found by singular value decomposition in time $\mathcal{O}(\min(n^2d, nd^2))$. Feldman et al. [2006] provided strong coresets with size exponential in $d$, $j$, and logarithmic in $n$ with $\mathcal{O}(n)$ construction time. The results were later improved to linear dependence on $d$, at the cost of $\mathcal{O}(njd)$ construction time [Feldman et al., 2010]. Feldman et al. [2013] proved that the sum of Euclidean distances from $n$ rows of $A$ to any $j$-dimensional subspace can be approximated up to factor $(1 + \varepsilon)$, with an additive constant which is the sum of a few last singular values of $A$, by projecting the points on the first $\mathcal{O}(j/\varepsilon)$ right singular vectors of $A$. Cohen et al. [2015] improved the latter result by a constant factor by projecting to a set of (random) orthonormal vectors approximating the right singular vectors of $A$.

NNMF and Latent Dirichlet Allocation (LDA) are constrained versions of the $j$-subspace problem. In NNMF the desired $j$-subspace must be spanned by positive vectors, and LDA is a generalization of NNMF, where we are given additional prior for every candidate solution. An important practical advantage of strong coresets (unlike weak coresets) is that they can be used with existing algorithms and heuristics for such constrained optimization problems.

**Integer $(k, j)$-projective clustering**. Varadarajan and Xiao [2012] analyze the *integer projective clustering* problem in which the points are on an integer grid and have their coordinates bounded by some polynomial in $n$. They proved that coresets sublinear in $n$ (but exponential in $j$) exist. Feldman et al. [2013] remove the dependence on $n$ at the cost of size being $poly(2^{kj}, \varepsilon^{-1})$. The idea is to first perform a deterministic dimensionality by projecting $A$ to its first $\mathcal{O}(j/\varepsilon)$ right singular vectors and apply the existing coreset constructions in the transformed space. Projecting the data to a set of (random) orthonor-mal vectors approximating the right singular vectors of *A*, Pratap and Sen [2016] compute coresets of the same size as in Feldman et al. [2013], albeit faster for small values of *d* and *n*.

## 3.3   Maximum likelihood estimation in mixture models

In hard clustering each data point is assigned to exactly one cluster. In contrast, in soft clustering each data point is assigned to each cluster with a certain probability. A prototypical example of soft clustering is fitting the parameters of a Gaussian mixture model in which one assumes that all data points are generated from a mixture of a finite number of Gaussians with unknown parameters. Other popular models include the Poisson mixture model, the mixture of multinomials and the mixture of exponentials.

Maximum likelihood estimation is a popular method of obtaining point estimates for various parametric models. Assuming the data was generated i.i.d. from some distribution $P(x \mid \theta)$, the negative log-likelihood of the data is

$$\mathcal{L}(\mathcal{X} \mid \theta) = - \sum_{x \in \mathcal{X}} \ln P(x \mid \theta).$$

The maximum likelihood estimate is obtained by solving

$$\theta^* = \arg\min_{\theta \in \mathcal{C}} \mathcal{L}(\mathcal{X} \mid \theta).$$

Finding the maximum likelihood estimate of a set of parameters for a single exponential family can be done efficiently trough the use of sufficient statistics. However, fitting the parameters of a mixture is a notoriously hard task – the mixture is not in the exponential family. The main idea behind constructing a coreset $\mathcal{C}$ is to reduce the problem of fitting a model on $\mathcal{X}$ to one of fitting a model on $\mathcal{C}$, since the optimal solution $\theta_{\mathcal{C}}$ is a good approximation (in terms of log-likelihood) of $\theta^*$. While finding the optimal $\theta_{\mathcal{C}}$ is a difficult problem, one can use a (weighted) variant of the EM algorithm to find a good solution. Moreover, if $|\mathcal{C}| \ll |\mathcal{X}|$, running EM on $\mathcal{C}$ is orders of magnitude faster than solving it on $\mathcal{X}$.

### 3.3.1 Gaussian Mixture Models

Arguably the most popular and flexible mixture model, the Gaussian mixture model, is defined as follows. Let $\theta = [(w_1, \mu_1, \Sigma_1), \ldots, (w_k, \mu_k, \Sigma_k)]$, and consider the distribution

$$P(x \mid \theta) = \sum_{i=1}^{k} w_i \mathcal{N}(x; \mu_i, \Sigma_i)$$

where $w_1, \ldots, w_k \geq 0$ are the mixture weights and $\sum_i w_i = 1$. Mean $\mu_i \in \mathbb{R}^d$ and covariance $\Sigma_i \in \mathbb{R}^{d \times d}$ parameterize the $i$-th mixture component, which is modeled as a multivariate normal distribution

$$\mathcal{N}(x; \mu_i, \Sigma_i) = \frac{1}{\sqrt{|2\pi\Sigma_i|}} \exp\left(-\frac{1}{2}(x - \mu_i)^T \Sigma_i^{-1}(x - \mu_i)\right).$$

Define $\mathfrak{C}$ to be the set of all mixtures of $k$ Gaussians $\theta$, such that all the eigenvalues of the covariance matrices of $\theta$ are bounded between $\lambda$ and $1/\lambda$ for $\lambda \in (0, 1)$ and consider models $\theta \in \mathfrak{C}$. Based on the idea studied in Arora and Kannan [2005], the negative log-likelihood can be decomposed as

$$\mathcal{L}(\mathcal{X} \mid \theta) = -\sum_{j=1}^{n} \ln \sum_{i=1}^{k} \frac{w_i}{\sqrt{|2\pi\Sigma_i|}} \exp\left(-\frac{1}{2}(x_j - \mu_i)^T \Sigma_i^{-1}(x_j - \mu_i)\right)$$
$$= -n \ln Z(\theta) + \phi(\mathcal{X} \mid \theta)$$

where function $\phi$ is defined as

$$\phi(\mathcal{X} \mid \theta) = -\sum_{j=1}^{n} \ln \sum_{i=1}^{k} \frac{w_i}{Z(\theta)\sqrt{|2\pi\Sigma_i|}} \exp\left(-\frac{1}{2}(x_j - \mu_i)^T \Sigma_i^{-1}(x_j - \mu_i)\right)$$

and $Z(\theta) = \sum_i w_i / \sqrt{|2\pi\Sigma_i|}$ is a normalizer which can be computed *exactly*, independently of the set $\mathcal{X}$. Critically, the function $\phi(\mathcal{X} \mid \theta)$ captures all dependencies of $\mathcal{L}(\mathcal{X} \mid \theta)$ on $\mathcal{X}$.

**Definition 3.1.** Let $\mathcal{X} \subseteq \mathbb{R}^d$, $k \in \mathbb{N}$, and $\varepsilon > 0$. $\mathcal{C}$ is a $(k, \varepsilon)$-*coreset* for $\mathcal{X}$, if for all mixtures $\theta \in \mathfrak{C}$ of $k$ Gaussians it holds that

$$(1 - \varepsilon)\phi(\mathcal{X} \mid \theta) \leq \phi(\mathcal{C} \mid \theta) \leq \phi(\mathcal{X} \mid \theta)(1 + \varepsilon).$$

Hence, $\mathcal{L}(\mathcal{C}_\varepsilon \mid \theta)$ uniformly approximates $\mathcal{L}(\mathcal{X} \mid \theta)$ (over $\theta \in \mathfrak{C}$) as $\varepsilon \to 0$. In fact, for a sequence of $(k,\varepsilon)$-coresets $\mathcal{C}_\varepsilon$ we have that

$$\sup_{\theta \in \mathfrak{C}} |\mathcal{L}(\mathcal{C}_\varepsilon \mid \theta) - \mathcal{L}(\mathcal{X} \mid \theta)| = \sup_{\theta \in \mathfrak{C}} |\phi(\mathcal{C}_\varepsilon \mid \theta) - \phi(\mathcal{X} \mid \theta)| \to 0.$$

Based on the framework presented in Chapter 2, the sensitivity can be bounded by $\mathcal{O}(k\lambda^{-4})$ and the pseudo-dimension of the associated function family is $\mathcal{O}(k^4 d^4)$ yielding a coreset of size $\Theta(d^4 k^6 \varepsilon^{-2} \lambda^{-8})$[Feldman et al., 2011, Lucic et al., 2017]. The coreset construction time is $\mathcal{O}(nkd)$.

### 3.3.2 Regular Exponential Family Mixtures

Banerjee et al. [2005] show that there is a bijection between regular exponential family distributions and Bregman divergences. In particular, the log-likelihood of exponential family mixture models can be expressed in terms of Bregman divergences. By considering the resulting objective function, one obtains *Bregman soft clustering*. More formally, let $d_\phi$ be a Bregman divergence and $\mathcal{X} \subseteq \mathcal{K}$ be a set of $n$ points. Let $k \in \mathbb{N}$, $w = (w_1, \ldots, w_k) \subseteq \mathbb{R}^k$ and $\theta = (\theta_1, \ldots, \theta_k) \subset \mathbb{R}^{kd}$ and let $Q$ be the concatenation of $w$ and $\theta$. The goal of Bregman soft clustering is to minimize

$$\text{cost}_{\text{s}}(\mathcal{X}, Q) = -\sum_{i=1}^n \ln \left( \sum_{j=1}^k w_j \exp\left(-d_\phi(x_i, \theta_j)\right) \right)$$

with respect to $Q$ under the constraint that $w_j > 0$, $1 \leq j \leq k$ and $\sum_{j=1}^k w_j = 1$. Now, consider a set $\tilde{\mathcal{X}}$ of $n$ points drawn independently from a stochastic source that is a mixture of $k$ densities of the same exponential family. The maximum likelihood estimation problem is

$$\max_{\{w_j, \theta_j\}_{j=1}^k} \mathcal{L}(\tilde{\mathcal{X}} \mid \theta) = \sum_{i=1}^n \ln\left(\sum_{j=1}^k w_j p_\psi(\tilde{x}_i \mid \theta_j)\right). \tag{3.1}$$

As shown by Banerjee et al. [2005] there is a bijection between regular exponential families and regular Bregman divergences that allows us

to rewrite (3.1) as

$$\mathcal{L}(\mathcal{X} \mid Q) = \sum_{i=1}^{n} \ln\left(\sum_{j=1}^{k} w_j \exp\left(-d_\phi(x, \eta_j)\right) b_\phi(x)\right)$$

$$= \sum_{i=1}^{n} \ln(b_\phi(x_i)) + \sum_{i=1}^{n} \ln\left(\sum_{j=1}^{k} w_j \exp\left(-d_\phi(x_i, \eta_j)\right)\right)$$

where $d_\phi$ is the corresponding Bregman divergence and both $\tilde{x}$ and $\eta_j$ are related to $x$ and $\theta_j$ by Legendre duality via $\phi$. Since the first summand is independent of the model parameters, maximizing the likelihood of the mixture model is equivalent to minimizing

$$\text{cost}_\text{S}(\mathcal{X}, Q) = -\sum_{i=1}^{n} \ln\left(\sum_{j=1}^{k} w_j \exp\left(-d_\phi(x_i, \theta_j)\right)\right).$$

Moreover, as $\varepsilon \to 0$

$$|\mathcal{L}(\mathcal{X}|Q) - \mathcal{L}(\mathcal{C}|Q)| = |\text{cost}_\text{S}(\mathcal{X}, Q) - \text{cost}_\text{S}(\mathcal{C}, Q)| \to 0$$

uniformly over $Q \in \mathcal{Q}$.

**Definition 3.2** ($\mu$-similar Bregman divergence). A Bregman divergence $d_\phi$ on domain $\mathcal{K} \subseteq \mathbb{R}^d$ is $\mu$-similar for some $\mu > 0$ iff there exists a positive definite matrix $A \in \mathbb{R}^{d \times d}$ such that, for each $p, q \in \mathcal{K}$,

$$\mu\, d_A(p, q) \leq d_\phi(p, q) \leq d_A(p, q)$$

where $d_A$ denotes the squared Mahalanobis distance.

Lucic et al. [2016b] show that the sensitivity for $\mu$-similar Bregman divergences can be bounded by $\mathcal{O}(k\mu^{-1})$ and the pseudo-dimension of the associated function family by $\mathcal{O}(k^4 d^2)$ yielding a coreset size of $\Theta\left(\frac{k^2 d^4 + k^2 \log(1/\delta)}{\mu^2 \varepsilon^2}\right)$. In fact, it is enough to compute the bicriteria using $D^2$ sampling with respect to the corresponding Mahalanobis distance. The coreset construction time is $\mathcal{O}(nkd)$.

## 3.4 Nonparametric mixture models

One of the main drawback of parametric mixture models is finding the "correct" number of clusters. Bayesian nonparametric generalizations of finite mixture models can be used to simultaneously estimate the number of components in a mixture model and the parameters of the individual mixture components. In essence, Bayesian nonparametric mixtures use mixing distributions consisting of a countably infinite number of atoms. Nevertheless, when applied to a finite training set, only a finite number of components will be used to model the data. The key idea is that the number of components can increase as the amount of samples increases. Orbanz and Teh [2011] provide an excellent review on the topic.

**Dirichlet Process Gaussian Mixture Model.** The Dirichlet process Gaussian mixture model is the generalization of the Gaussian mixture model where the number of components is countably infinite. The means $\mu_i$ are sampled from a Dirichlet process with a concentration parameter $\alpha$ and a spherical Gaussian with variance $\rho$ as its base distribution. Cluster assignments of $n$ data points are then sampled using a *Chinese Restaurant Process* with concentration parameter $\alpha$. Each data point is finally generated from a Gaussian with the assigned cluster center as its mean and variance $\sigma$. By analyzing the case where $\sigma$ approaches zero, Kulis and Jordan [2012] show that the Gibbs sampler converges to a hard clustering algorithm. The resulting cost function is also know as *DP-Means*

$$\text{cost}(\mathcal{P}, Q) = \sum_{p \in \mathcal{P}} w_p \min_{q \in Q} \text{dist}(p, q)^2 + |Q|\lambda$$

where $\lambda > 0$, $\mathcal{P}$ is a weighted set of $n$ points in $\mathbb{R}^d$ and $Q \subset \mathbb{R}^d$ a non-empty set of cluster centers. This objective function resembles the $k$-means objective where one allows infinitely many clusters, but the addition of a new cluster is penalized by $\lambda$.

Bachem et al. [2015] first show the existence of $\varepsilon$-coresets for the DP-Means clustering problem that are sublinear in the number of data points $n$ if the optimal number of centers $k^*$ is sublinear in $n$. Formally, they show the existence of an $\varepsilon$-coreset of size $\mathcal{O}\left(d^d k^* \log n \varepsilon^{-d}\right)$ where $k^*$ is the optimal number of centers. Then, they show a practical algorithm with a polynomial dependence on the relevant quantities. The main contribution is the a $\mathcal{O}(\log_2 k')$ competitive bicriteria approximation based on $D^2$-sampling which infers the number of centers $k'$ from the data using a stopping condition. Using this bicriteria and a sensi-tivity based importance sampling scheme they obtain coresets of size $\mathcal{O}\Big((dk'^3 \log k' + k'^2 \log \frac{1}{\delta})\varepsilon^{-2}\Big)$.

### 3.4.1 Classification and regression

**Regression.** Boutsidis et al. [2012] provide a deterministic, polynomial-time algorithm for constructing a weak coreset for arbitrarily constrained linear regression. The coreset is of size $\mathcal{O}(k/\varepsilon^2)$, where $k$ is the rank ("effective dimension") of the data matrix. The results extend to multiple response regression setting. The authors provide (almost) matching lower-bounds and improve the $\mathcal{O}(k \log k/\varepsilon^2)$ results of Drineas et al. [2006], Boutsidis and Drineas [2009]. The results are based on sparsification tools from linear algebra Boutsidis et al. [2014]. Recently, Huggins et al. [2016] have shown that the upper bound on sensitivity computed by $k$-clustering is useful in the context of Bayesian Logistic Regression.

Now consider the case of two-class classification of a set of points $\mathcal{X} = \mathcal{X}_+ \cup \mathcal{X}_-$. In Support Vector Machines (SVM) our goal is to find a hyperplane $h$ such that $\mathcal{X}_+$ and $\mathcal{X}_-$ are on opposite sides of $h$ and that the minimum distance between $h$ and the points of $\mathcal{X}$ (the margin) is maximized.

**Minimum enclosing radius ball and SVM**. The minimum enclosing ball MEB$(\mathcal{X})$ is the smallest ball containing $\mathcal{X}$. Tsang et al. [2005] show that many kernel methods can be equivalently formulated as MEB problems. The exact methods for the MEB problem are not efficient for $d > 30$. Hence, approximate solutions to the MEB problems were considered. In a breakthrough result, Badoiu and Clarkson [2002] show the existence of weak coresets of size $\lceil 1/\varepsilon \rceil$ and prove that it is tight in the worst case. A weak coreset coreset guarantees that the radius smallest ball containing $\mathcal{C}$ is within $1 + \varepsilon$ of the smallest ball containing $\mathcal{X}$ (for the strong coreset consult Agarwal et al. [2004]). The exceptional feature of the coreset construction algorithm is that the number of iterations is only $\mathcal{O}(\varepsilon^{-2})$ and does not depend on $d$ and $n$. The latter bound was improved to $\mathcal{O}(\varepsilon^{-1})$ by Kumar et al. [2003].

Based on the coreset construction algorithm in Badoiu and Clarkson [2002], Tsang et al. [2005] propose Core Vector Machine algorithm

which can be used with nonlinear kernels and has a time complexity that is linear in $n$ and a space complexity that is independent of $n$. The algorithm provides an $(1+\varepsilon)^2$ approximation to the SVM objective. The algorithm can be further simplified by using $(1+\varepsilon)$ approximation algorithms for enclosing balls with fixed radius which leads to good empirical results Tsang et al. [2007].

**Communication efficient ERM.** Reddi et al. [2015] introduce a general strategy for designing communication efficient empirical loss minimization algorithms. The main insight is that at each iteration of the optimization procedure there exists only a small set of critical points - the ones currently close to the margin. Moreover, for SVMs, due to the piecewise nature of hinge loss, points far away from the margin can be represented by a linear function. As such, the coreset consists of a few points close to the margin and the linear functional representing far away points. The authors demonstrate good empirical results on both SVM and logistic regression. We note that the notion of a coreset is different: it is not a weighted subset of the original data set, and the guarantee is additive.

## 3.5 Streaming and distributed settings

One advantage of coresets is that they can be constructed in parallel, as well as in a streaming setting where data points arrive one by one, and it is impossible to remember the entire data set due to memory constraints. The key insight is that coresets satisfy certain composition properties, which have previously been used by Har-Peled and Mazumdar [2004] for streaming and parallel construction of coresets for geometric clustering problems such as $k$-median and $k$-means.

1. Let $\mathcal{C}_1$ be a $(k,\varepsilon)$-coreset for $\mathcal{X}_1$, and $\mathcal{C}_2$ be a $(k,\varepsilon)$-coreset for $\mathcal{X}_2$. Then $\mathcal{C}_1 \cup \mathcal{C}_2$ is a $(k,\varepsilon)$-coreset for $\mathcal{X}_1 \cup \mathcal{X}_2$.

2. Let $\mathcal{C}$ be a $(k,\varepsilon)$-coreset for $\mathcal{X}$, and $\mathcal{C}'$ be a $(k,\delta)$-coreset for $\mathcal{C}$. Then $\mathcal{C}'$ is a $(k,(1+\varepsilon)(1+\delta)-1)$-coreset for $\mathcal{X}$.

**Streaming construction.** In the streaming setting we have limited memory and are allowed one pass trough the data. The goal is to pro-duce a coreset of the points seen with update time per point as well as the total memory bounded by some polynomial in $\log n$. The composition properties of strong coresets can again be leveraged by applying the *merge-reduce* framework [Chazelle and Matoušek, 1996]. The idea is to construct and store in memory a coreset for every block of consecutive points arriving in a stream. As soon as we have two coresets in memory we merge them which results in a $(k, \varepsilon)$-coreset via property (1). Then we can compress them by computing a single coreset from the merged coresets via property (2) to avoid the increase in the coreset size. In order to maintain a balanced binary tree we assign a *level* to each coreset. Initially the level is set to zero, and is increased by one every time a compression operation is executed. Provided that all merging operations are performed on the coresets of the same level guarantees that the resulting computation tree has height at most $\log n$ as depicted in Figure 3.1. Hence, for a predefined error of $\varepsilon$ it suffices that each coreset on the first level is a $(k, \varepsilon/\log n)$-coreset. A subtle issue arises in the streaming setting as we do not know $n$ a priori. Intuitively, this issue can be resolved by computing coresets for data batches of exponentially increasing size – the total time and space requirements are dominated by the last batch whose size is upper bounded by $n$. We refer the reader to Feldman et al. [2013] for details. This dependency of the error on $\log n$ can sometimes be removed by careful analysis and delayed merging and compression steps [Chazelle and Matoušek, 1996].

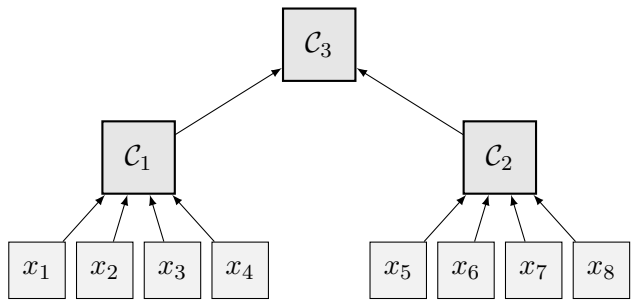

**Figure 3.1:** Tree construction for generating coresets in parallel or from data streams. Black arrows indicate "merge-and-compress" operations. The (intermediate) coresets $\mathcal{C}_1$ and $\mathcal{C}_2$ are enumerated in the order in which they would be generated in the streaming case. In the parallel case, $\mathcal{C}_1$ and $\mathcal{C}_2$ would be constructed in parallel, followed by parallel construction of $\mathcal{C}_3$.

**Distributed construction.** In this setting the data set $\mathcal{X}$ is partitioned between $m$ machines and our goal is to compute a coreset for $\mathcal{X}$. Property (1) immediately suggests a simple algorithm: construct a $(k, \varepsilon)$-coreset on each machine and merge the obtained coresets. This resulting coreset is guaranteed to be a $(k, \varepsilon)$-coreset for $\mathcal{X}$.